\documentclass[10pt,twocolumn,letterpaper]{article}

\usepackage{cvpr}
\usepackage{times}
\usepackage{epsfig}
\usepackage{graphicx}
\usepackage{amsmath}
\usepackage{amssymb}

\usepackage{url}

\usepackage{wrapfig}

\usepackage{epsfig}
\usepackage{pifont}
\usepackage{amsbsy}
\usepackage{graphicx}
\usepackage{amsmath}
\usepackage{amssymb}
\newcommand{\xmark}{\ding{55}}%
\usepackage{bbding}
\usepackage{mathrsfs}
\usepackage{multirow}

  \usepackage{flushend}
\usepackage[breaklinks=true,bookmarks=false]{hyperref}

\cvprfinalcopy % *** Uncomment this line for the final submission

\def\cvprPaperID{6786} % *** Enter the CVPR Paper ID here

\ifcvprfinal\fi
	
\begin{document}

%%%%%%%%% TITLE
% \title{AFHN: A Few-Shot Learning Approach \\ with Diverse and Discriminative Feature Synthesis}

% \title{Diversity-Regularized Adversarial Augmentation Network for Few-Shot Image Classification}
% Diversity and discriminability Regularized Adversarial Augmentation Network for few-shot learning

\title{Adversarial Feature Hallucination Networks for Few-Shot Learning}
% \title{On Data Augmentation for Few-Shot Learning: An Adversarial Learning Approach Unifying Diversity and Discriminability}

% \author{Kai Li$^{1}$, Martin Renqiang Min$^{2}$, Yun Fu$^{1,3}$ \\
% $^1$Department of Electrical and Computer Engineering, Northeastern University, Boston, USA \\
% $^2$NEC Laboratories America \\
% $^3$Khoury College of Computer Science, Northeastern University, Boston, USA \\
% {\tt\small kaili@ece.neu.edu, renqiang@nec-labs.com, yunfu@ece.neu.edu}

\author{Kai Li$^{1}$, Yulun Zhang$^{1}$, Kunpeng Li$^{1}$, Yun Fu$^{1,2}$ \\
$^1$Department of Electrical and Computer Engineering, Northeastern University, Boston, USA \\
$^2$Khoury College of Computer Science, Northeastern University, Boston, USA \\
{\tt\small \{kaili,kunpengli,yunfu\}@ece.neu.edu, yulun100@gmail.com}
% \and
}

\maketitle
\thispagestyle{empty}

%\thispagestyle{empty}
% Rich and accessible labeled data trigger for the recent flourish of deep learning in various tasks.
%%%%%%%%% ABSTRACT
\begin{abstract}
The recent flourish of deep learning in various tasks is largely accredited to the rich and accessible labeled data. Nonetheless, massive supervision remains a luxury for many real applications, boosting great interest in label-scarce techniques such as few-shot learning (FSL), which aims to learn concept of new classes with a few labeled samples. A natural approach to FSL is data augmentation and many recent works have proved the feasibility by proposing various data synthesis models. However, these models fail to well secure the discriminability and diversity of the synthesized data and thus often produce undesirable results. In this paper, we propose Adversarial Feature Hallucination Networks (AFHN) which is based on conditional Wasserstein Generative Adversarial networks (cWGAN) and hallucinates diverse and discriminative features conditioned on the few labeled samples. Two novel regularizers, i.e., the classification regularizer and the anti-collapse regularizer, are incorporated into AFHN to encourage discriminability and diversity of the synthesized features, respectively. Ablation study verifies the effectiveness of the proposed cWGAN based feature hallucination framework and the proposed regularizers. Comparative results on three common benchmark datasets substantiate the superiority of AFHN to existing data augmentation based FSL approaches and other state-of-the-art ones. 
\end{abstract}

\section{Introduction}
The rich and accessible labeled data fuel the revolutionary success of deep learning \cite{deng2009imagenet,zhao2017multi,li2018support}.  However, in many specific real applications, only limited labeled data are available. This motivates the investigation of few-shot learning (FSL) where we need to learn concept of new classes based on a few labeled samples. To combat with deficiency of labeled data, some FSL methods resort to enhance the discriminability of the feature representations such that a simple linear classifier learned from a few labeled samples can reach satisfactory classification results \cite{vinyals2016matching,snell2017prototypical,triantafillou2017few}. Another category of methods investigate techniques of quickly and effectively updating a deep neural network with a few labeled data, either by learning a meta-network and the corresponding updating rules \cite{finn2017model,li2017meta,ravi2016optimization,munkhdalai2017meta}, or by learning a meta-learner model that generates some components of a classification network directly from the labeled samples \cite{li2019novel,gidaris2018dynamic,rusu2018meta}.  Alternatively, the third group of methods address this problem with data augmentation by distorting the labeled images or synthesizing new images/features based on the labeled ones \cite{Chen2019Image,gao2018low,schwartz2018delta,chen2018semantic}.

Our proposed method falls into the data augmentation based category. The basic assumption of approaches in this category is that the intra-class cross-sample relationship learned from seen (training) classes can be applied to unseen (test) classes. Once the cross-sample relationship is modeled and learned from seen classes, it can be applied on the few labeled samples of unseen class to hallucinated new ones. It is believed that the augmented samples can diversify the intra-class variance and thus help reach sharper classification boundaries \cite{zhang2018metagan}. Whatever data augmentation technique is used, it is critical to secure discriminability of the augmented samples, as otherwise they shall cast catastrophic impact on the classifier. On the other hand, the decision boundary of a classifier can be determined precisely only when labeled samples exhibit sufficient intra-class variance. Thus, diversity of the augmented samples is also of a crucial role. This is in fact the essential motivation of investigating data augmentation for FSL, as a few labeled samples encapsulate limited intra-class variance.

Though various data augmentation based FSL methods have been proposed recently, they fail to simultaneously guarantee discriminability and diversity of the synthesized samples. Some methods learn a finite set of transformation mappings between samples in each base (label-rich) classes and directly apply them to seed samples of novel (label-scarce) classes. However, the arbitrary mapping may destroy discriminability of the synthesized samples \cite{Chen2019multi,hariharan2017low,schwartz2018delta}. Other methods synthesize samples specifically for certain tasks which regularize the synthesis process \cite{wang2018low,munkhdalai2017meta}. Thus, these methods can guarantee discriminability of the synthesized samples. But the task would constrain the synthesis process and consequently the synthesized samples tend to collapse into certain modes, thus failing to secure diversity.

To avoid limitations of the existing methods, we propose Adversarial Feature Hallucination Networks (AFHN) which consists of a novel conditional Wasserstein Generative Adversarial Networks (cWGAN) \cite{gulrajani2017improved} based feature synthesis framework and two novel regularizers. Unlike many other data augmentation based FSL approaches that perform data augmentation in the image space \cite{Chen2019ImageAAAI,Chen2019multi,Chen2019Image}, our cWGAN based framework hallucinates new features by using the features of the seed labeled samples as the conditional context. To secure discriminability of the synthesized features, AFHN incorporates a novel classification regularizer that constrains \textit{the synthesized features being of high correlation with features of real samples from the same class while of low correlation with those from the different classes}. With this constraint, the generator is encouraged to generate features encapsulating discriminative information of the class used as the conditional context.

It is more complicated to ensure diversity of the synthesized features, as conditional GANs are notoriously susceptible to the mode collapse problem that only samples from limited distribution modes are synthesized. This is caused by the use of usually high dimensional and structured data as the condition tends to make the generator ignore the latent code, which controls diversity. To avoid this problem, we propose a novel anti-collapse regularizer which assigns high penalty for the case where mode collapse likely occurs. It is derived from the observation that \textit{noise vectors that are closer in the latent code space are more likely to be collapsed into the same mode when mapped to the feature space}. We directly penalize the ratio of the dissimilarity of the two synthesized feature vectors and the dissimilarity of the two noise vectors generating them. With this constraint, the generator is forced to explore minor distribution modes, thus encouraging diversity of the synthesized features.

With discriminative and diverse features synthesized, we can get highly effective classifiers and accordingly appealing recognition results. In summary, the contributions of this paper are as follows: (1) We propose a novel cWGAN based FSL framework which synthesizes fake features by taking those of the few labeled samples as the conditional context. (2) We propose two novel regularizers that guarantee discriminability and diversity of the synthesized features. (3) The proposed method reaches the state-of-the-art performance on three common benchmark datasets.

\section{Related Work}
Regarding the perspective of addressing FSL, existing algorithms can generally be divided into three categories. The first category of methods aim to enhance the discriminability of the feature representations extracted from images. To this goal, a number of methods resort to deep metric learning and learn deep embedding models that produce discriminative feature for any given image \cite{ren2018meta,vinyals2016matching,snell2017prototypical,triantafillou2017few}. The difference lies in the loss functions used. Other methods following this line focus on improving the deep metric learning results by learning a separate similarity metric network \cite{yang2018learning}, task dependent adaptive metric \cite{oreshkin2018tadam}, patch-wise similarity weighted metric \cite{hao2019collect}, neural graph based metric \cite{kim2019edge,liu2018learning}, etc.

A more common category of algorithms address FSL by enhancing  flexibility of a model such that it can be readily updated using a few labeled samples. These methods utilize meta-learning, also called learning to learn, which learns an algorithm (meta-learner) that outputs a model (the learner) that can be applied on a new task when given some information (meta-data) about that task. Following this line, some approaches aim to optimize a meta-learned classification model such that it can be  easily fine-tuned using a few labeled data \cite{ravi2016optimization,finn2017model,li2017meta,li2017meta,ravi2016optimization,munkhdalai2017meta,nichol2018first}. Other approaches adopt neural network generation and train a meta-learning network which can adaptively generate entire or some components of a classification neural network from a few labeled samples of novel classes \cite{qiao2017few,gidaris2018dynamic,li2019rethinking,li2019novel}. The generated neural network is supposed to be more effective to classify unlabeled samples from the novel classes, as it is generated from the labeled samples and encapsulates discriminative information about these classes.

The last category of methods combat deficiency of the labeled data directly with data augmentation. Some methods try to employ additional samples by some forms of transfer learning from external data \cite{ren2018meta,wang2016learning}. More popular approaches perform data augmentation internally by applying transformations on the labeled images or the corresponding feature representations. Naively distorting the images with common transformation techniques (e.g., adding Gaussian perturbation, color jittering, etc.) is particularly risky as it likely jeopardizes the discriminative content in the images. This is undesirable for FSL as we only have a very limited number of images to be utilized; quality control of the synthesizing results for any single image is crucial as otherwise the classifier could be ruined by the low-quality images.  Chen et al. propose a series of methods of performing quality-controlled image distortions by applying perturbation in the semantic feature space \cite{Chen2019multi}, shuffling image patches \cite{Chen2019ImageAAAI} and explicitly learning an image transformation network \cite{Chen2019Image}. Performing data augmentation in the feature space seems more promising as the feature variance directly affects the classifier. Many approaches with this idea have been proposed by hallucinating new samples for novel class based on seen classes \cite{schwartz2018delta,hariharan2017low}, composing synthesized representations \cite{chen2018semantic,Yu2018Low}, and using GANs \cite{gao2018low,zhang2018metagan}.

This paper proposes Adversarial Feature Hallucination Networks (AFHN), a new GAN-based FSL model that augments labeled samples by synthesizing fake features conditioned on those of the labeled ones. AFHN significantly differs from the two existing GAN based models \cite{zhang2018metagan,gao2018low} in the following aspects. First, AFHN builds upon Wasserstein GAN (WGAN) model which is known for more stable performance, while \cite{zhang2018metagan,gao2018low} adopt the conventional GAN framework. Second, neither \cite{zhang2018metagan} nor \cite{gao2018low} has a classification regularizer. The most similar optimization objective in \cite{gao2018low} is the one which optimizes the synthesized features as the outlier class (relative to the real class), while that in \cite{zhang2018metagan} is a cycle-consistency objective. We instead regularize the synthesized features of being high correlation with real features from the same classes and low correlation with those from the different classes. Third, After training the generator, we learn a standard Softmax classifier using the synthesize features, while \cite{zhang2018metagan,gao2018low} utilize them to enhance existing FSL methods. Last, we further propose the novel anti-collapse regularizer to encourage diversity of synthesized features, while \cite{zhang2018metagan,gao2018low} do not. 

AFHN also bears some similarity with an existing feature hallucination based FSL method \cite{wang2018low}. But apparently we adopt the GAN framework which has the discriminator to regularize the features produced by the generator, while \cite{wang2018low} uses the simple generative model. Besides, AFHN synthesizes new features to learn a standard Softmax classifier for new classes, while \cite{wang2018low} utilizes them to enhance existing FSL classifier. Moreover, we aim to hallucinate diverse features with the novel anti-collapse regularizer, while \cite{wang2018low} does not have such an objective.

\section{Algorithm}
In this section, we first briefly introduce Wasserstein GAN and then elaborate the details of how we build the proposed AFHN model upon it.

\subsection{Wasserstein GAN}
GAN is a recently proposed generative model that has shown impressive performance on synthesizing realistic images. The generative process in GAN is modeled as a game between two  competitive models, the generator and the discriminator. The generator aims to generate from noise fake samples as realistic as possible such that the discriminator cannot tell whether they are real or fake. The discriminator instead tries the best to make the correct judgment. This adversarial game pushes the generator to extensively explores the data distribution and consequently produces more visually appealing samples than conventional generative models. However, it is known that GAN is highly unstable in training.
\cite{arjovsky2017wasserstein} analyzes the convergence properties of the objective function of GAN and proposes the Wasserstein GAN (WGAN) which utilizes the Wasserstein distance in the objective function and is shown of better theoretical properties than the vanilla GAN.
We adopt the improved variant of WGAN \cite{gulrajani2017improved}, which optimizes the following min-max problem,
\begin{equation}
\begin{array}{cl}
\underset{G}{\min} \hspace{1pt} \underset{D}{\max} & \underset{\tilde{\mathbf{x}}\sim\mathbb{P}_g}{\mathbb{E}}[D(\tilde{\mathbf{x}})] 
- \underset{\mathbf{x}\sim\mathbb{P}_r}{\mathbb{E}} [D(\textbf{x})] \\
& +  \lambda \underset{\hat{\mathbf{x}}\sim\mathbb{P}_{\hat{\mathbf{x}}}}{\mathbb{E}}[(\|\nabla_{\hat{\mathbf{x}}} D(\hat{\mathbf{x}})\|_2-1)^2],
\end{array}
\label{wgan}
% \label{cls_loss}
\end{equation}
where $\mathbb{P}_r$ is the data distribution and $\mathbb{P}_g$ is the model distribution defined by $\tilde{\mathbf{x}}\sim G(\mathbf{z})$, with $\mathbf{z}\sim p(\mathbf{z})$ randomly sampled from noise distribution $p$. $\mathbb{P}_{\hat{\mathbf{x}}}$ is defined by sampling uniformly along straight lines between pairs of points sampled from the data distribution $\mathbb{P}_r$ and the generator distribution $\mathbb{P}_g$, i.e., $\hat{\mathbf{x}} = \alpha \mathbf{x}+(1-\alpha)\tilde{\mathbf{x}}$ with $\alpha \sim U (0, 1)$. The first two terms approximate the Wasserstein distance and the third term penalizes the gradient norm of $\hat{\mathbf{x}}$.

\subsection{Adversarial Feature Hallucination Networks}
Following the literature, we formally define FSL as follows: Given a distribution of tasks $P(\mathcal{T})$, a sample task $\mathcal{T}$$\sim$$P(\mathcal{T})$ 
is a tuple $\mathcal{T}=(S_\mathcal{T}, Q_\mathcal{T})$ where the support set 
$S_\mathcal{T}=\{\{\textbf{x}_{i, j}\}^{K}_{i=1}, y_j\}^{N}_{j=1}$  
contains $K$ labeled samples from each of the $N$ classes. This is usually known as $K$-shot $N$-way
classification. 
$Q_\mathcal{T}=\{(\textbf{x}_q, y_q)\}^Q_{q=1}$ is the query set where the samples come from the same $N$ classes as the support set $S_\mathcal{T}$. The learning objective is to minimize the classification prediction risk of $Q_\mathcal{T}$, according to $S_\mathcal{T}$. 

\begin{figure*}
  \begin{center}
    \includegraphics[width=1.0\textwidth]{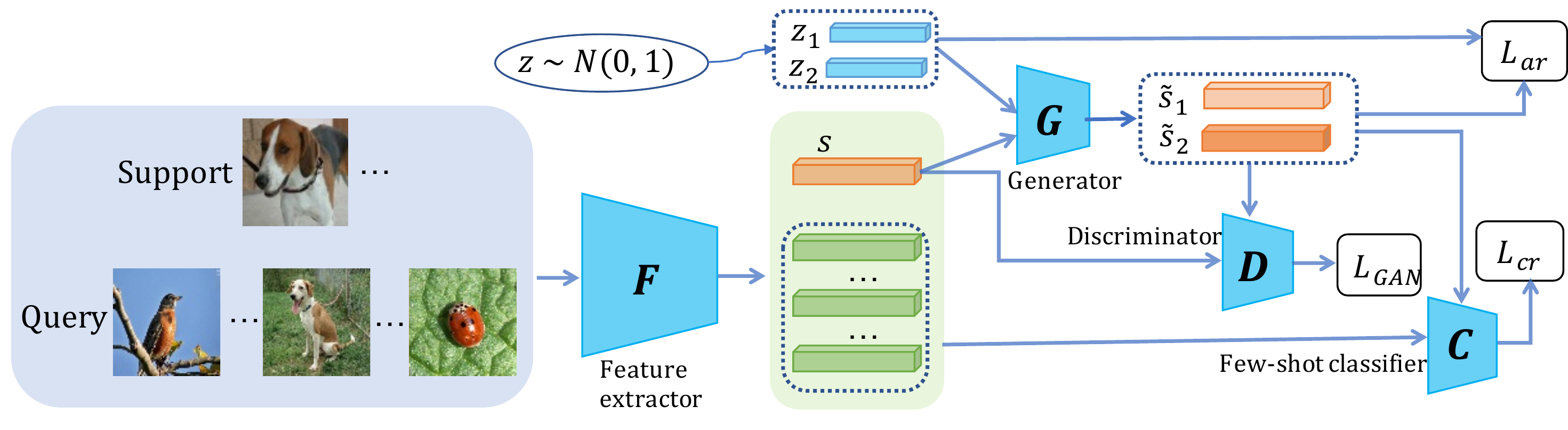}         
  \end{center}
% \vspace{-10pt}
\caption{Framework of the proposed AFHN. 
AFHN takes as input a support set and a query set where images in the query set belongs to the sampled classes in the support set. Each image in the support set is fed to the feature extraction network $F$, resulting the feature embedding $\textbf{s}$. With $\textbf{s}$, feature generator $G$ synthesizes two fake features $\tilde{\textbf{s}}_1$ and $\tilde{\textbf{s}}_2$, by combining $\textbf{s}$ with two randomly sampled variables $\textbf{z}_1$ and $\textbf{z}_2$. Discriminator $D$ discriminates real feature $\textbf{s}$ and fake features $\tilde{\textbf{s}}_1$ and $\tilde{\textbf{s}}_2$, resulting in the GAN loss $L_{GAN}$. By analyzing the relationship between ($\textbf{z}_1$, $\textbf{z}_2$) and ($\tilde{\textbf{s}}_1$, $\tilde{\textbf{s}}_2$), we get the anti-collapse loss $L_{ar}$. The proposed few-shot classifier classifies the features of the query images based on the fake features $\tilde{\textbf{s}}_1$ and $\tilde{\textbf{s}}_2$. This results in the classification loss $L_{cr}$.}
\label{fremework}   
% \vspace{-5pt}
\end{figure*}

The proposed AFHN approaches this problem by proposing a general conditional WGAN based FSL framework and two novel regularization terms. Figure~\ref{fremework} illustrates the training pipeline.

\noindent\textbf{FSL framework with conditional WGAN}. 
For a typical FSL task $\mathcal{T}=(S_\mathcal{T}, Q_\mathcal{T})$,
the feature extraction network $F$ produces a representation vector for each image.
Specifically for an 
image from the support set $(\textbf{x}, y)\in S_\mathcal{T}$, $F$ generates
\begin{equation}
\mathbf{s} = F(\mathbf{x}).
\end{equation}
When there are multiple samples for class $y$, i.e., $K > 1$, we simply average the feature vectors and take the averaged vector as the prototype of class $y$ \cite{snell2017prototypical}. Conditioned on $\mathbf{s}$, we synthesize fake features for the class. 

% For a typical $K$-shot $N$-way
%\underset{\mathbf{x}\sim\mathbb{P}_r}{\mathbb{E}}

Unlike previous GAN models which sample a single random noise variable from some distribution, we sample two noise variables $\textbf{z}_1$ and $\textbf{z}_2\sim N(0, 1)$.  The generator $G$ synthesizes fake feature $\tilde{\textbf{s}}_1$ ($\tilde{\textbf{s}}_2$) taking as input $\textbf{z}_1$ ($\textbf{z}_2$) and the class prototype $\mathbf{s}$,
\begin{equation}
\tilde{\textbf{s}}_i = G(\textbf{s}, \textbf{z}_i), \hspace{8pt} \\\ i = 1, 2.
\label{generator}
\end{equation}
The generator $G$ aims to synthesize $\tilde{\textbf{s}}_i$ to be as similar as possible to $\textbf{s}$.
% on the other hand, aim to fool $D$ to make the correct decision.% The generator, on the other hand, aim to fool $D$ to make the correct decision. 
The discriminator $D$ tries to discern $\tilde{\textbf{s}}_i$ as fake and $\textbf{s}$ as real. 
Within the WGAN framework, the adversarial training objective is as follows,
\begin{equation}
\begin{array}{cl}
L_{GAN_i}=\underset{(\textbf{x},y)\sim S_\mathcal{T}}{\mathbb{E}}[D(\tilde{\mathbf{s}}_i, \mathbf{s})] 
- \underset{(\textbf{x},y)\sim S_\mathcal{T}}{\mathbb{E}} [D(\textbf{s}, \mathbf{s})]  \\
 + \lambda \underset{(\textbf{x},y)\sim S_\mathcal{T}}{\mathbb{E}}[(\|\nabla_{\hat{\mathbf{s}}_i} D(\hat{\mathbf{s}}_i, \mathbf{s})\|_2-1)^2], 
\hspace{3pt} \hspace{3pt}  i = 1, 2. 
\end{array}
\label{cwgan}
%\hspace{1pt} \textrm{for} \\\ i = 1, 2. 
\end{equation}

Simply training the model with the above GAN loss does not guarantee the generated features are well suited for learning a discriminative classifier because it neglects the inter-class competing information among different classes. Moreover, since the conditioned feature vectors are of high dimension and structured, it is likely that the generator will neglect the noise vectors and all synthesized features collapse to a single or few points in the feature space, i.e., the so-called mode collapse problem. To avoid these problems, we append the objective function with a classification regularization term and an anti-collapse regularization term, aiming to encourage both diversity and discriminability of the synthesized features.

\noindent\textbf{Classification regularizer}.
As our training objective is to classify well samples in the query set $Q_\mathcal{T}$, given the support set $S_\mathcal{T}$, we encourage discriminability of the synthesized features by requiring them serving well the classification task as the real features. 
Inspired by \cite{snell2017prototypical}, we define a non-parametric FSL classifier which calculates the possibility of a query image $(\textbf{x}_q, y_q)\in Q_\mathcal{T}$ of being the same class as synthesized feature $\tilde{\textbf{s}}_i$ as
% $\textbf{x}$ as
\begin{equation}
P(y_q=y|\textbf{x}_q) = \frac{\exp(\cos(\tilde{\mathbf{s}}_i, \mathbf{q}))}{\sum_{j=1}^N \exp(\cos(\tilde{\mathbf{s}}^j_i, \textbf{q}))},
\label{probability} 
\end{equation}
where $\textbf{q}=F(\textbf{x}_q)$. $\tilde{\mathbf{s}}^j_i$ is the synthesized feature for the $j$-th class and $\cos(\textbf{a}, \textbf{b})$ is the Cosine similarity of two vectors.
The adoption of Cosine similarity, rather than Euclidean distance as in \cite{snell2017prototypical}, is inspired by a recent FSL algorithm \cite{gidaris2018dynamic} which proves that Cosine similarity can bound and reduce variance of the features and result in models of better generalization. 

With the proposed FSL classifier, the classification regularizer in a typical FSL task is defined as follows:
\begin{equation}
%\begin{array}{cl}
L_{cr_i}=\underset{{\tiny(\mathbf{x}_q, y_q)\sim Q_{\mathcal{T}}}}{\mathbb{E}}\Big[\frac{1}{N}\sum^N_{y=1}y\log [-P(y_q=y|\mathbf{x}_q)]\Big], 
%\end{array}
\end{equation}
for $i=1,2$.  We can see that this regularizer explicitly encourages the synthesized features to have high correlation with features from the same class (the conditional context), while low correlation with features from the different classes. To achieve this, the synthesized features must encapsulate discriminative information about the conditioned class and thus secure discriminability. 
%This thus ensures feature discriminability. 

\noindent\textbf{Anti-collapse regularizer}. 
GAN models are known for suffering from the notorious mode collapse problem, especially conditional GANs where structured and high-dimensional data (e.g., images) are usually used as the conditional contexts. As a consequence, the generator likely ignores the latent code (noises) that accounts for diversity and focuses only on the conditional contexts, which is undesirable. Specifically to our case, our goal is to augment the few labeled samples in the feature space; when mode collapse occurs, all synthesized features may collapse to a single or a few points in the feature space, failing to diversify the labeled samples. Observing that noise vectors that are closer in the latent code space are more likely to be collapsed into the same mode when mapped to the feature space, we directly penalize the ratio of the dissimilarity two synthesized feature vectors and the dissimilarity of the two noise vectors generating them.

Remember that we sample two random variables $\textbf{z}_1$ and $\textbf{z}_2$. We generate two fake feature vectors $\tilde{\textbf{s}}_1$ and $\tilde{\textbf{s}}_2$ from them.
%$\tilde{\textbf{s}}_1=G(\textbf{s}, \textbf{z}_1)$ and $\tilde{\textbf{s}}_2=G(\textbf{s}, \textbf{z}_2)$. 
When $\textbf{z}_1$ and $\textbf{z}_2$ are closer, $\textbf{s}_1$ and $\textbf{s}_2$ are more likely to be collapsed into the same mode.
To mitigate this, we define the anti-collapse regularization term as
\begin{equation}
\mathcal{L}_{ar} = \underset{(\mathbf{x}, y)\sim S_{\mathcal{T}}}{\mathbb{E}}\Big[ \frac{1-\cos(\tilde{\mathbf{s}}_1, \tilde{\mathbf{s}}_2)}{1-\cos(\mathbf{z}_1, \mathbf{z}_2)}\Big].
\label{reg_loss}
\end{equation}
We can observe that this term amplifies the dissimilarity of
the two fake feature vectors when the latent codes generating them are of high similarity. With the case mode collapse more likely occurs being assigned with higher penalty, the generator is forced to mine minor modes in the feature space during training. The discriminator will also handle fake features from the minor modes. Thus, it is expected that more diverse features can be synthesized when applying the generator on novel classes.

With the above two regularization terms, we reach our final training objective as 
\begin{equation}
\underset{G}{\min} \hspace{1pt} \underset{D}{\max} \hspace{2pt} \hspace{1pt} \sum^2_{i=1} L_{GAN_i} + \alpha \sum^2_{i=1}L_{cr_i} 
+ \beta \frac{1}{L_{ar}},
\label{Obj}
\end{equation}
where $\alpha$ and $\beta$ are two hyper-parameters. \textbf{Algorithm 1} outlines the main training steps of the proposed method.

\begin{table}[t]
  \small
  \centering
  \begin{tabular}{l}
    \hline
    \noindent \textbf{Algorithm 1.} Proposed FSL algorithm  \\\hline 
    \textbf{Input:} Training set $\mathcal{D}_t=\{\mathcal{X}_t, \mathcal{Y}_t\}$, parameters $\lambda$, $\alpha$, and $\beta$. \\ 
    \textbf{Output:} Feature extractor $F$, generator $G$, discriminator $D$. \\\hline
    1. Train $F$ as a standard classification task using $\mathcal{D}_t$. \\     
    \textbf{while} not done \textbf{do}\\
    \hspace{3mm} // \textit{Fix $G$ and update $D$}. \\
    \hspace{3mm} 2. Sample from $\mathcal{D}_t$ a batch of FSL tasks $\mathcal{T}^d_i\sim p(\mathcal{D}_t)$.\\
    \hspace{3mm} \textbf{For} each $\mathcal{T}^d_i$ \textbf{do} \\ 
    \hspace{6mm} 3. Sample a support set $S_\mathcal{T}=\{\{\textbf{x}_{i, j}\}^{K}_{i=1}, y_j\}^{N}_{j=1}$ and \\ 
    \hspace{10mm} query set $Q_\mathcal{T}=\{\{\textbf{x}_{k, j}\}^{Q}_{k=1}, y_j\}^{N}_{j=1}$.  \\    
    \hspace{6mm} 4. Compute prototypes of the $N$ classes $\mathcal{P}=\{\textbf{s}_j\}^{N}_{j=1}$, \\ 
    \hspace{10mm}  where $\mathbf{s}_j = \frac{1}{K} \sum_{i=1}^{K} F(\textbf{x}_{i, j})$. \\
    \hspace{6mm} 5. Sample $N$ noise variables $\mathcal{Z}_1=\{\textbf{z}^j_1\}_{j=1}^N$ and \\ 
    \hspace{10mm}	variables $\mathcal{Z}_2=\{\textbf{z}^j_2\}_{j=1}^N$. \\ 
    \hspace{6mm} 6. Generate fake feature sets $\tilde{\mathcal{Z}}_1=\{\tilde{\textbf{z}}^j_1\}_{j=1}^N$ \\
    \hspace{10mm}    and $\tilde{\mathcal{Z}}_2=\{\tilde{\textbf{z}}^j_2\}_{j=1}^N$ according to Eq. \eqref{generator}. \\ 
    \hspace{6mm} 7. Update $D$ by maximizing Eq. \eqref{Obj}. \\
    \hspace{3mm} \textbf{end For} \\\vspace{1pt}
    \hspace{3mm} // \textit{Fix $D$ and update $G$}. \\
    \hspace{3mm} 8. Sample from $\mathcal{D}_t$ a batch of FSL tasks $\mathcal{T}^g_i\sim p(\mathcal{D}_t)$. \\
    \hspace{3mm} \textbf{For} each $\mathcal{T}^g_i$ \textbf{do} \\ 
    \hspace{6mm} 9. Execute steps 3 - 7.				\\
    \hspace{6mm} 10. Update $G$ by minimizing Eq. \eqref{Obj}. \\
    \hspace{3mm} \textbf{end For} \\ 
    \textbf{end while} 
   \\ \hline 
   \vspace{0.5pt}
  \end{tabular}
  \vspace{-15pt}
\end{table}

\subsection{Classification with Synthesized Samples}
In the test stage, given an FSL task $\mathcal{T}'=(S'_\mathcal{T}, Q'_\mathcal{T})$ randomly sampled from the test set that the classes have no overlap with those in the training set, we first augment the labeled support set $S'_\mathcal{T}$ with the learned generator $G$. Then, we train a classifier with the augmented supported set. The classifier is used to classify samples from the query set $Q'_\mathcal{T}$.
Specifically, suppose after data augmentation, we get an enlarged support set
$\hat{S}'_\mathcal{R} = \{(\textbf{s}_1, y_1 ), (\textbf{s}_2 , y_2), \cdots, (\textbf{s}_{N\times K'} , y_{N\times K'}\}$  
where $K'$ is the number of samples synthesized for each class. 
With $\hat{S}'_\mathcal{R}$, we train a standard Softmax classifier $\textbf{f}_c$ as 
\begin{equation}
\min_{\theta} \underset{(\mathbf{s}, y)\sim \hat{S}'_\mathcal{R}}{\mathbb{E}}\log [-P(y|\mathbf{s}; \theta)],
\label{cls_loss}
\end{equation}
where $\theta$ is the parameter of $\textbf{f}_c$. With $\textbf{f}_c$, we classify samples from $Q'_\mathcal{T}$.

\section{Experiments}
We evaluate AFHN on three common benchmark datasets, namely, \textit{Mini-ImageNet} \cite{vinyals2016matching}, \textit{CUB} \cite{wah2011multiclass} and \textit{CIFAR100} \cite{krizhevsky2009learning}. 
The \textit{Mini-ImageNet} dataset is a subset of ImageNet. It has 60,000 images from 100 classes, 600 images for each class. We follow previous methods and use the splits in \cite{ravi2016optimization} for evaluation, i.e., 64, 16, 20 classes as training, validation, and testing sets, respectively.  The \textit{CUB} dataset is a fine-grained dataset of totally 11,788 images from 200 categories of birds. We use the split in \cite{hilliard2018few} and 100, 50, 50 classes for training, validation, and testing, respectively. The CIFAR-100 dataset contains 60,000 images from 100 categories. We use the same data split as in \cite{zhou2018deep}. In particular, 64, 16 and 20 classes are used for training, validation and testing, respectively. 

Following previous methods, we evaluate 5-way 1-shot and 5-way 5-shot classification tasks where each task instance involves classifying test images from 5 sampled classes with 1 or 5 randomly sampled images for each class as the support set. In order to reduce variance, we repeat the evaluation task 600 times and report the mean of the accuracy with a 95\% confidence interval.

\subsection{Implementation Details}
Following the previous data augmentation based methods \cite{schwartz2018delta,Chen2019multi,Chen2019Image}, we use ResNet18 \cite{he2016deep} as our feature extraction network $F$.
We implement the generator $G$ as a two-layer MLP, with LeakyReLU activation for the first layer and ReLU activation for the second one. The dimension of the hidden layer is 1024. The discriminator is also a two-layer MLP, with LeakyReLU as the activation function for the first layer. The dimension of the hidden layer is also 1024. The noise vectors $\textbf{z}_1$ and $\textbf{z}_2$ are drawn from a unit Gaussian with the same dimensionality as the feature embeddings. 

Following the data augmentation based FSL methods \cite{schwartz2018delta,Chen2019multi}, we perform two-step training procedures. In the first step, we only train the feature extraction network $F$ as a multi-class classification task using only the training split. We use Adam optimizer with an initial learning rate $10^{-3}$ which decays to the half every 10 epochs. We train $F$ with 100 epochs with batch size of 128. In the second training stage, we train the generator and discriminator alternatively, using features extracted by $F$ 
and update $G$ after every 5 updates of $D$. We also use Adam optimizer which has an initial learning rate of $10^{-5}$ and decays to the half every 20 epochs for both $G$ and $D$. We train the whole network with 100 epochs with 600 randomly sampled FSL tasks in each epoch.  For the hyper-parameters, we set $\lambda=10$ as suggested by \cite{gulrajani2017improved}, and $\alpha=\beta=1$ for all the three datasets. During the test stage, we synthesize 300 fake features for each class.

The code is developed based on PyTorch.

\begin{table}
\small
\renewcommand{\tabcolsep}{5pt}
\begin{center}
\begin{tabular}{|l|ccccc|}\hline
cWGAN              & \xmark    & \xmark                    & \Checkmark  & \Checkmark   & \Checkmark \\       
CR                & \xmark    & \Checkmark                    & \xmark      & \Checkmark   & \Checkmark \\
AR                & \xmark    & \xmark        & \xmark      & \xmark             & \Checkmark \\ \hline
          			& 52.73   & 55.65           & 57.58       &  60.56            &  62.38 \\ \hline
\end{tabular}
\end{center}
\vspace{-5pt}
\caption{Ablation study on the \textit{Mini-ImageNet} dataset for the 5-way 1-shot setting.  cWGAN, CR, and AR represent the conditional WGAN framework, the classification regularizer, and the anti-collapse regularizer, respectively. 
The baseline result (52.73) is obtained by applying the SVM classifier directly on ResNet18 features without data augmentation. The result (55.65) with only CR added is obtained from the synthesized features produced by the generator without the discriminator and AR during training.}
\label{Table_ablation}
\end{table}

\begin{table*}
\small
\renewcommand{\tabcolsep}{10pt}
 \begin{center}
\begin{tabular}{|l|l|c|c|c|c|} \hline   
& & Backbone & Reference & 1-shot & 5-shot  \\ \hline\hline
& ResNet18 + SVM (baseline)
& ResNet18
&
& 52.73$\pm$1.44 
& 73.31$\pm$0.81
\\ \hline
\multirow{10}{*}{MetricL}
& Matching Net \cite{vinyals2016matching}
& Conv-64F
& NeurIPS'16
& 43.56$\pm$0.84
& 55.31$\pm$0.73
\\ 
& PROTO Net \cite{snell2017prototypical}
& Conv-64F
& NeurIPS'17
& 49.42$\pm$0.78
& 68.20$\pm$0.66
\\ 
& MM-Net \cite{cai2018memory}
& Conv-64F
& CVPR'18
& 53.37$\pm$0.48 
& 66.97$\pm$0.35
\\
& GNN \cite{garcia2017few}
& Conv-256F
& Arxiv'17
& 50.33$\pm$0.36
& 66.41$\pm$0.63
\\
& RELATION NET \cite{yang2018learning}
& Conv-64F
& CVPR'18
& 50.44$\pm$0.82
& 65.32$\pm$0.70
\\
& DN4 \cite{li2019revisiting}
& Conv-64F
& CVPR'19
& 51.24$\pm$0.74
& 71.02$\pm$0.64
\\
& TPN \cite{liu2018learning}
& ResNet8
& ICLR'19
& 55.51$\pm$0.86
& 69.86$\pm$0.65
\\
& PARN \cite{wu2019parn}
& Conv-64F
& ICCV'19
& 55.22$\pm$0.84
& 71.55$\pm$0.66
\\
& SAML \cite{hao2019collect}
& Conv-64F
& ICCV'19
& 57.69$\pm$0.20
& 73.03$\pm$0.16
\\
& DCEM \cite{dvornik2019diversity}
& ResNet18
& ICCV'19
& 58.71$\pm$0.62 
& 77.28$\pm$0.46
\\ \hline\hline
\multirow{9}{*}{MetaL}
& MAML \cite{finn2017model}
& Conv-32F
& ICML'17
& 48.70$\pm$1.84
& 63.11$\pm$0.92
\\ 
& META-LSTM \cite{ravi2016optimization}
& Conv-32F
& ICLR'17
& 43.44$\pm$0.77
& 60.60$\pm$0.71
\\ 
& SNAIL \cite{mishra2017simple}
& ResNet-256F
& ICLR'18
& 55.71$\pm$0.99
& 68.88$\pm$0.92
\\
& MACO \cite{hilliard2018few}
& Conv-32F
& Arxiv'18
& 41.09$\pm$0.32
& 58.32$\pm$0.21
\\
& DFSVL \cite{gidaris2018dynamic}
& Conv-64F
& CVPR'18
&55.95$\pm$0.89
&73.00$\pm$0.68
\\
& META-SGD \cite{li2017meta}
& Conv-32F
& Arxiv'17
& 50.47$\pm$1.87
& 64.03$\pm$0.94
\\
& PPA \cite{qiao2017few}
& WRN-28-10
& CVPR'18
& 59.60$\pm$0.41 
& 73.74$\pm$0.19 
\\
& UFDA \cite{li2019novel}
& ResNet18
& CIKM'19
& 60.51
& 77.08
\\
& LEO \cite{rusu2018meta}
& WRN-28-10
& ICLR'19
& 61.76$\pm$0.08
& 77.59$\pm$0.12
\\ \hline\hline
\multirow{5}{*}{DataAug}
& MetaGAN \cite{zhang2018metagan}
& Conv-32F
& NeurIPS'18
& 52.71$\pm$0.64
& 68.63$\pm$0.67
\\
& Dual TriNet \cite{Chen2019Image}
& ResNet18
& TIP'19
& 58.80$\pm$1.37
& 76.71$\pm$0.69
\\
& $\Delta$-encoder \cite{schwartz2018delta}
& ResNet18
& NeurIPS'18
& 59.90
& 69.70
\\ 
& IDeMe-Net \cite{Chen2019Image}
& ResNet18
& CVPR'19
& 59.14$\pm$0.86
& 74.63$\pm$0.74
\\ \cline{2-6}
& AFHN (Proposed)
& ResNet18
&
& \textbf{62.38$\pm$0.72}
& \textbf{78.16$\pm$0.56}
\\ \hline
\end{tabular} 
% \\\text{(a)}
% \end{minipage}
% \vspace{8pt}
\end{center}
 \vspace{-5pt}
\caption{Few-shot classification accuracy on \textit{Mini-Imagenet}. 
``MetricL'', ``MetaL'' and ``DataAug'' represent metric learning based category, meta-learning based category and data augmentation based category, respectively.
The “$\pm$” indicates 95\% confidence intervals over tasks. The best results are in \textbf{bold}.}
 \label{result_fsl_mini}
 \vspace{-5pt}
\end{table*}

\begin{table*}
	\small
	\renewcommand{\tabcolsep}{6pt}	
	\begin{center}
		\begin{tabular}{|l|l|c|c|c|c|c|c|} \hline   
			& & \multirow{2}{*}{Backbone} & \multirow{2}{*}{Reference} & \multicolumn{2}{c|}{\textit{CUB}}   & \multicolumn{2}{c|}{\textit{CIFAR100}}   \\ \cline{5-8}
			& & & & 1-shot & 5-shot & 1-shot & 5-shot    \\ \hline
			& ResNet18 + SVM (baseline)
			& ResNet18
			&
			& 66.54$\pm$0.53
			& 82.38$\pm$0.43
			& 59.65$\pm$0.78
			& 76.75$\pm$0.73
			\\ \hline\hline
			\multirow{4}{*}{MetricL}
			& Matching Net \cite{vinyals2016matching}
			& Conv-64F
			& NeurIPS'16
			& 49.34
			& 59.31
			& 50.53$\pm$0.87
			& 60.30$\pm$0.82
			\\ 
			& PROTO Net \cite{snell2017prototypical}
			& Conv-64F
			& NeurIPS'17
			& 45.27
			& 56.35
			& -
			& -
			\\
			& DN4 \cite{li2019revisiting}
			& Conv-64F
			& CVPR'19
			& 53.15$\pm$0.84
			& 81.90$\pm$0.60
			& -
			& -
			\\
			& SAML \cite{hao2019collect}
			& Conv-64F
			& ICCV'19
			& 69.33$\pm$0.22 
			& 81.56$\pm$0.15
			& -
			& -
			\\ \hline\hline
			\multirow{4}{*}{MetaL}
			& MAML \cite{finn2017model}
			& Conv-32F
			& ICML'17
			& 38.43
			& 59.15
			& 49.28$\pm$0.90
			& 58.30$\pm$0.80
			\\ 
			& META-LSTM \cite{ravi2016optimization}
			& Conv-32F
			& ICLR'17
			& 40.43
			& 49.65
			& -
			& -
			\\ 
			& MACO \cite{hilliard2018few}
			& Conv-32F
			& Arxiv'18
			& 60.76 
			& 74.96
			& -
			& -
			\\
			& META-SGD \cite{li2017meta}
			& Conv-32F
			& Arxiv'17
			& 66.90
			& 77.10
			& 61.60
			& 77.90
%			\\
%			& UFDA \cite{li2017meta}
%			& ResNet18
%			& CIKM'19
%			& 69.59
%			& \textbf{84.83}
%			& -
%			& -
			\\ \hline\hline
			\multirow{3}{*}{DataAug}
			& Dual TriNet \cite{Chen2019multi}
			& ResNet18
			& TIP'19
			& 69.61
			& 84.10
			& 63.41$\pm$0.64
			& 78.43$\pm$0.64
			\\
			& $\Delta$-encoder \cite{schwartz2018delta}
			& ResNet18
			& NeurIPS'18
			& 69.80$\pm$0.46  
			& 82.60$\pm$0.35
			& 66.70  
			& 79.80
			\\ \cline{2-8}	
			& AFHN  (Proposed)
			& ResNet18
			& 
			& \textbf{70.53}$\pm$1.01
			& 83.95$\pm$0.63 
			& \textbf{68.32$\pm$0.93}
			& \textbf{81.45$\pm$0.87}
			\\ \hline
		\end{tabular} 
		\vspace{-3pt}
	\end{center}
	\caption{Few-shot classification accuracy on \textit{CUB} and \textit{CIFAR100}. Please refer Table \ref{result_fsl_mini} for details.}
	\label{result_fsl_cub_cifar100}
%	\vspace{-5pt}
\end{table*}

\subsection{Ablation Study}
The proposed AFHN consists of the novel conditional WGAN (cWGAN) based feature synthesize framework and the two regularizers that encourage diversity and discriminability of the synthesized features, i.e., the Classification Regularier (CR) and Anti-collapse Regularizer (AR).  
To evaluate the effectiveness and impact of these components, we conduct ablation study on the \textit{Mini-ImageNet} dataset for the 5-way 1-shot setting.  The results are shown in Table \ref{Table_ablation}.

\begin{figure*}
	\begin{center}
		\begin{minipage}{.33\textwidth}
			\centering
			\includegraphics[width=1.\linewidth]{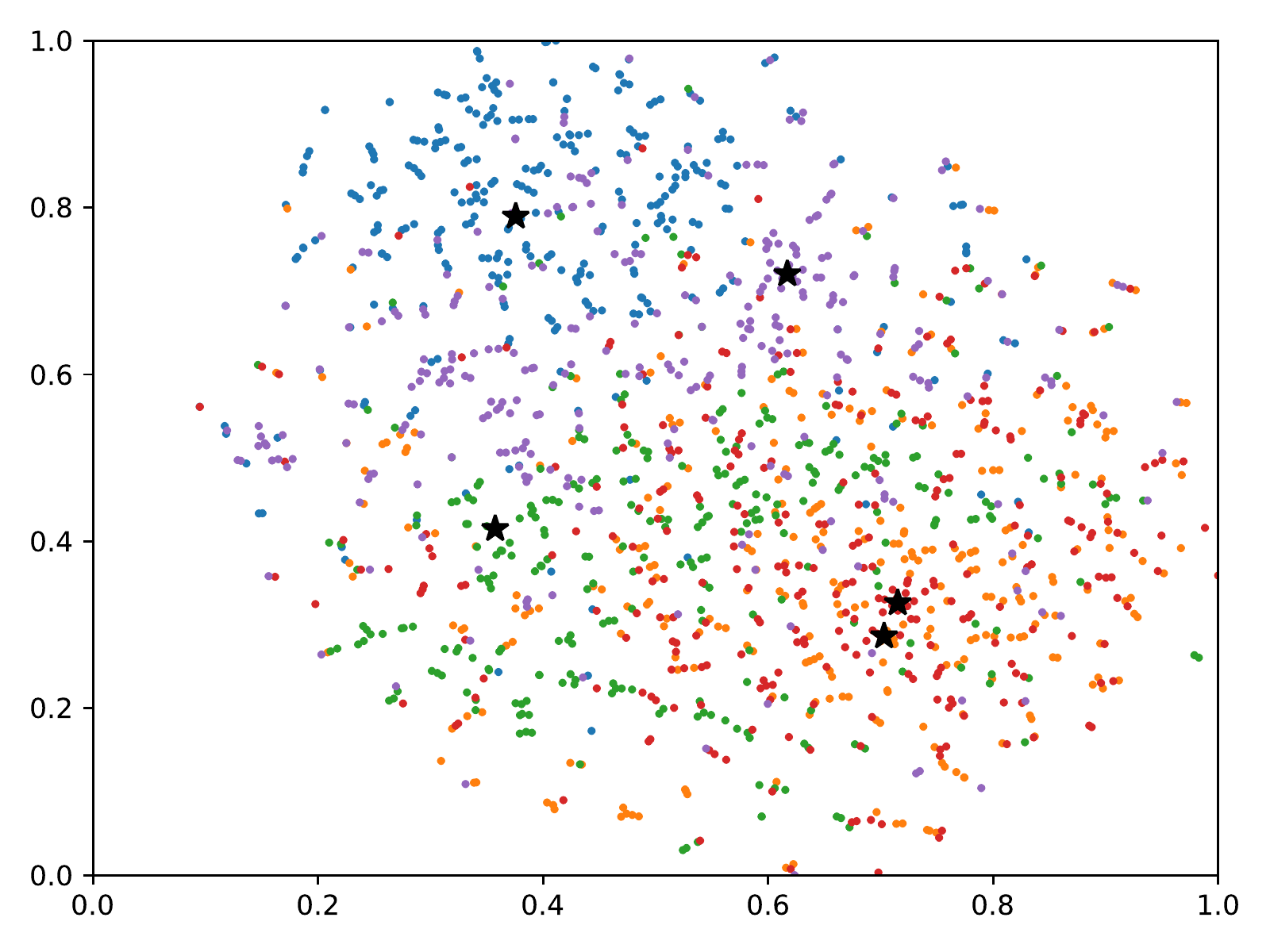}
		\end{minipage}%
		\begin{minipage}{0.33\textwidth}
			\centering
			\includegraphics[width=1.\linewidth]{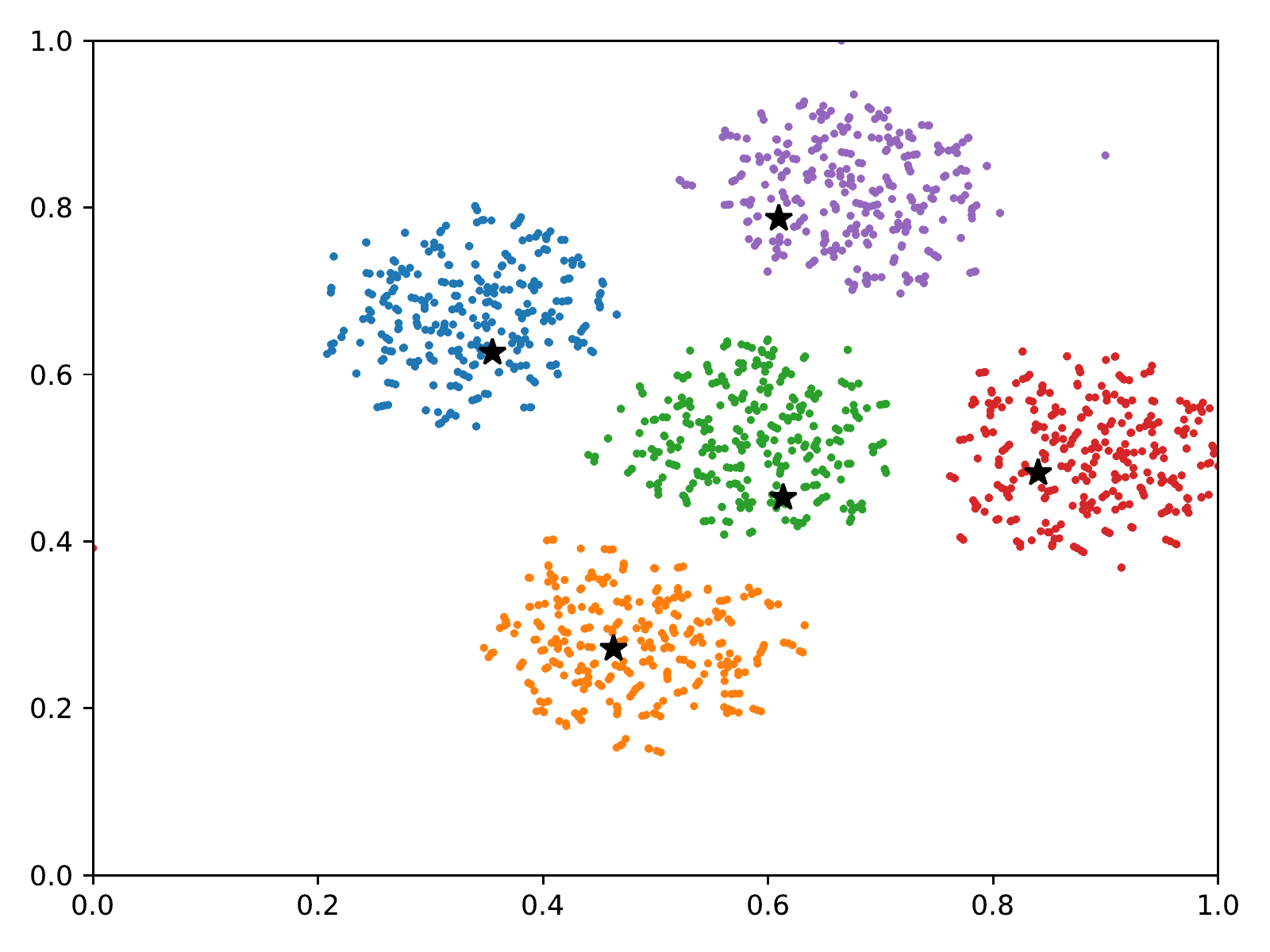}
		\end{minipage}
		\begin{minipage}{0.33\textwidth}
			\centering
			\includegraphics[width=1.\linewidth]{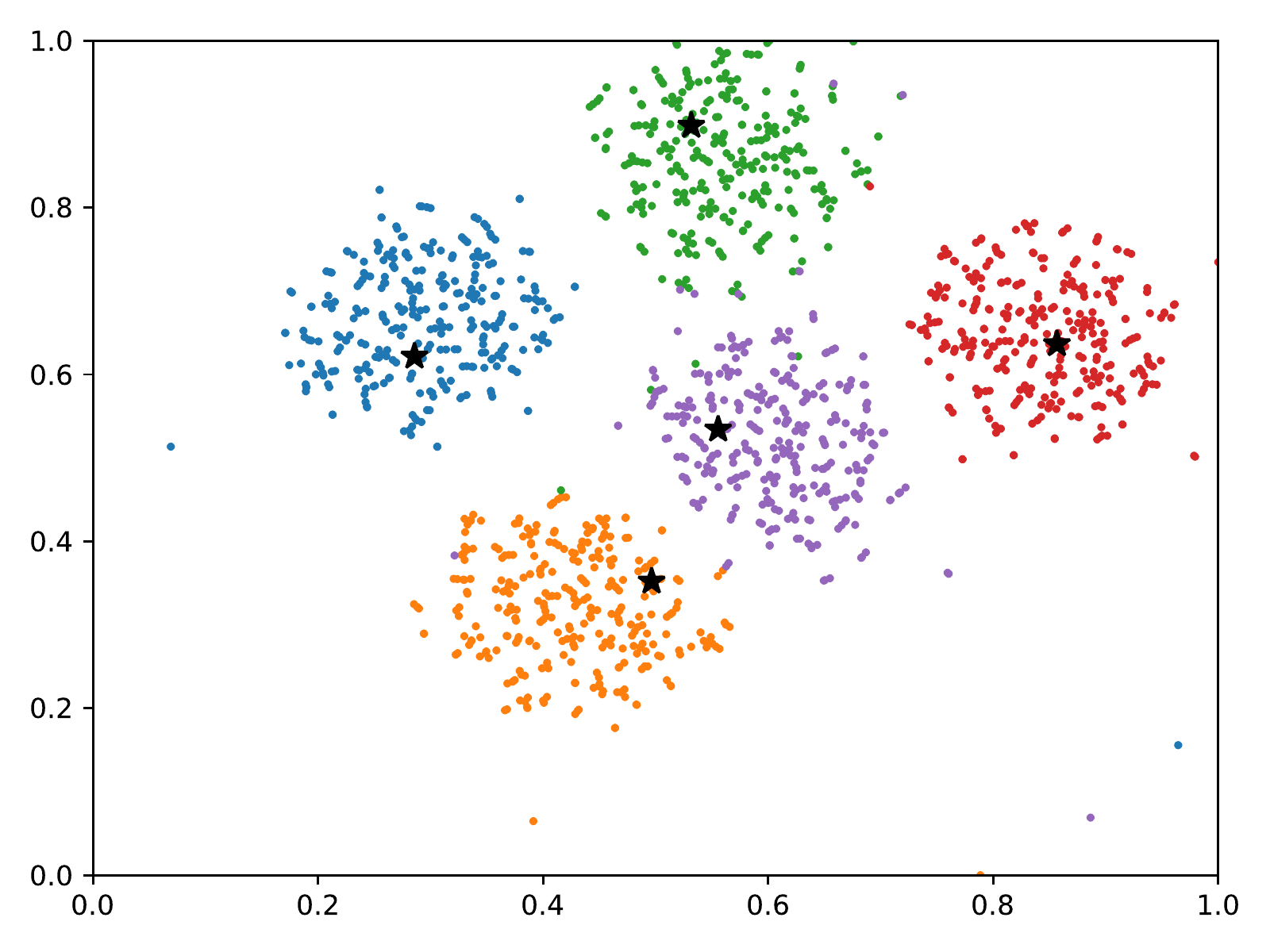}
		\end{minipage}  \\  
		
		\begin{minipage}{.33\textwidth}
			\centering
			\text{cWGAN}
		\end{minipage}%3
		\begin{minipage}{0.33\textwidth}
			\centering
			\text{cWGAN + CR}
		\end{minipage}
		\begin{minipage}{0.33\textwidth}
			\centering
			\text{cWGAN + CR + AR}
		\end{minipage}      
		\vspace{-8pt}
	\end{center}
	\caption{t-SNE \cite{maaten2008visualizing} visualization of synthesized feature embeddings. 
		The real features are indicated by $\star$. Different colors represent different classes.}
	\label{tsne}
	\vspace{-5pt}
\end{figure*}

\begin{figure}
	\begin{center}
		\includegraphics[width=0.3\textwidth]{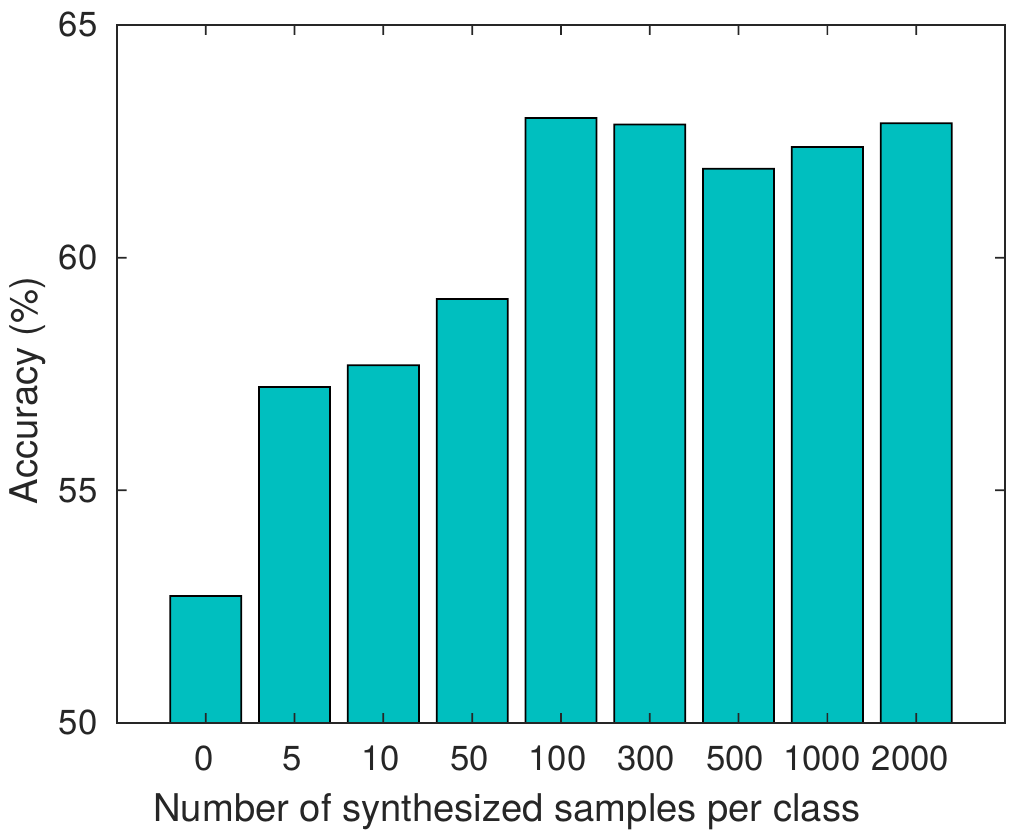}     
	\end{center}
	\vspace{-10pt}
	\caption{Impact of the number of synthesized samples for each class on the \textit{Mini-ImageNet} dataset.}
	\label{impact_of_sample_number}  
	\vspace{-8pt}
\end{figure}

% lie between the manifolds of different real data classes,
\noindent\textbf{CR}. This regularizer constrains the synthesized features to have desirable classification property such that we can train from them a discriminative classifier. We can see that when it is used as the only regularization for the generator, it raises the baseline result from 52.73 to 55.65. On the other hand, when it is used along with cWGAN (the discriminator regularizes the generated features, resulting in the GAN loss), it helps further boost the performance from 57.58 to 60.56. Therefore, in the both cases (with and without cWGAN), CR helps enhance discriminability of the synthesized features and leads to performance boost.

\noindent\textbf{cWGAN}. Compared with the baseline (without data augmentation), cWGAN helps raise the accuracy from 52.73 to 57.58.  This is because the synthesized features enhance the intra-class variance, which makes classification decision boundaries much sharper.  Moreover, with CR as the regularizer, our cWGAN based generative model boosts the performance of the naive generative model from 55.65 to 60.56. This further substantiates the effectiveness of the proposed cWGAN framework. The performance gain is due to the adversarial game between the generator and the discriminator, which enhances the generator's capability of modeling complex data distribution among training data. The enhanced generator is therefore able to synthesize features of both higher diversity and discriminability.

As mentioned in the related work, one of the major differences of the proposed AFHN from the other feature hallucination based FSL method \cite{wang2018low} is that AFHN is an adversarial generative model while \cite{wang2018low} uses a naive generative model. This study thus evidences the advantage of AFHN over \cite{wang2018low}.

\noindent\textbf{AR}. AR aims to encourage the diversity of the synthesized features by explicitly penalizing the case where mode collapses more likely occur. Table \ref{Table_ablation} shows that it further brings about 2\% performance gains, thus proving its effectiveness.

\subsection{Comparative Results}
\noindent\textbf{Mini-Imagenet}. 
\textit{Mini-Imagenet} is the most extensively evaluated dataset.  From Table \ref{result_fsl_mini} we can observe that AFHN attains the new state-of-the-art, for both the 1-shot and 5-shot setting. Compared with the other four data augmentation based methods, AFHN reaches significant improvements: it beats $\Delta$-encoder \cite{schwartz2018delta} by more than 8\% for the 5-shot setting and Dual TriNet \cite{Chen2019multi} by more than 3\% for the 1-shot setting.  Compared with MetaGAN \cite{zhang2018metagan} which is also based on GAN, AFHN achieves about 10\% improvements for both the 1-shot and 5-shot settings. Besides the significant advantages over the peer data augmentation based methods, AFHN also exhibits remarkable advantages over the other two categories of methods. It beats the best metric learning based method DCEM \cite{dvornik2019diversity} by about 3.5\% for the 1-shot setting. It also performs better than the state-of-the-art meta-learning based algorithms.
% Similarly, AFHN reaches about 3\% and 4\% gains relative to WRN \cite{qiao2017few}, the best meta-learning based method.
Compared with the baseline method, ``ResNet18+SVM'', AFHN reaches about 10\% and 5\% improvements for the 1-shot and 5-shot settings, respectively. This substantiates the effectiveness of our proposed data augmentation techniques.

\noindent\textbf{CUB}. This is a fine-grained bird dataset widely used for fine-grained classification. Recently, it has been employed for few-shot classification evaluation. Thus, relatively less results are reported on this dataset. From Table \ref{result_fsl_cub_cifar100} we can see that AFHN reaches comparable results with both the other two data augmentation based methods Dual TriNet and $\Delta$-encoder. It beats the best metric learning based method SAML \cite{hao2019collect} by 2.4\% for the 5-shot setting, and performs significantly better than the meta-learning based methods. Compared with the baseline, we only have a moderate improvement in the 1-shot setting and reach only a marginal boost for the 5-shot setting. We speculate the reason is that this dataset is relatively small, less than 60 images per class on average; a large number of classes only have about 30 images. Due to the small scale of this dataset, the intra-class variance is less significant than that of the \textit{Mini-Imagenet} dataset, such that 5 labeled samples are sufficient to capture most of the intra-class variance. Performing data augmentation is less crucial than that for the other datasets. 

\noindent\textbf{CIFAR100}. 
This dataset has the identical structure as the \textit{Mini-ImageNet} dataset.  Table \ref{result_fsl_cub_cifar100} shows that AFHN performs the best over all the existing methods and the advantages are sometimes significant. AFHN beats Dual TriNet by 5\% and 3\% for 1-shot and 5-shot respectively. Compared with the best meta-learning based method, we get 7\% and 4\% improvements for the 1-shot and 5-shot respectively. Compared with the baseline method, AFHN also reach remarkable gains. We reach about 10\% and 5\% improvements for 1-shot and 5-shot respectively. This great improvement convincingly substantiates the effectiveness of our GAN based data augmentation method for solving the FSL problem.

In summary, among all the three datasets, we reach significant improvements over existing state-of-the-art methods for two of them, while being comparable for the left one. For all the datasets, our method reaches significant boost to the baseline method where there is no data augmentation. These experiments substantiate the effectiveness and superiority of the proposed method.

\subsection{Further Analysis}
\noindent\textbf{Impact of the number of synthesized features}.
Figure \ref{impact_of_sample_number} shows the analysis on \textit{Mini-ImageNet} about the recognition accuracy with respect to the number of synthesized features for each class during test. 
We can observe that the classification accuracy keeps boosted with more features synthesized at the beginning, and remains stable with even more synthesized samples. 
This is reasonable because the class variance encapsulated by the few labeled samples has a upper bound; data augmentation based on these labeled samples can enlarge the variance to some extent, but it is still bounded by the few labeled samples themselves. When it reaches the peak, the performance reasonably turns stable.

\noindent\textbf{Visualization of synthesized features}.
We showed quantitatively in the ablation study that owing to the CR and AR regularizers, we can generate diverse and discriminative features which bring significant performance gains. Here we further study the effect of the two regularizers by showing the t-SNE visualization of the synthesized features. As shown in Figure \ref{tsne}, the synthesized features of different classes mix up together when using only cWGAN for augmentation. As analyzed before, cWGAN does not guarantee synthesizing semantically meaningful features. The problem is substantially resolved when we train cWGAN with CR. The synthesized features exhibit clear clustering structure, which helps train a discriminative classifier. Furthermore, with AR added, the synthesized features still exhibit favorable clustering structure. But taking a closer look of the visualization, we can find that the features synthesized with AR added are more diverse than that without it: the clusterings are less compact, stretched to larger regions, and even contains some noises. This shows AR indeed helps diversify the synthesized features.

\section{Conclusions}
We introduce the Adversarial Feature Hallucination Networks (AFHN), a new data augmentation based few-shot learning approach. AFHN consists of a novel conditional Wasserstein GAN (cWGAN) based feature synthesis framework, the classification regularizer (CR) and the anti-collapse regularizer (AR). 
Based on cWGAN, our framework synthesizes fake features for new classes by using the features of the few labeled samples as the conditional context. CR secures feature discriminability by requiring the synthesized features to be of high similarity with features of the samples from the same classes, while of low similarity with those from the different classes. AR aims to enhance the diversity of the synthesized features by directly penalizing the cases where the mode collapse problem likely occurs. The ablation study shows the effectiveness of the cWGAN based feature synthesis framework, as well as the two regularizers. Comparative results verify the superiority of AFHN to the existing data augmentation based FSL approaches as well as other state-of-the-art ones. 

\vspace{5pt}
\noindent\textbf{Acknowledgement}: This research is supported by the U.S. Army Research Office Award W911NF-17-1-0367.

\clearpage
{\small
\bibliographystyle{ieee_fullname}
\bibliography{egbib}
}

\end{document}